\documentclass{article}

\usepackage{arxiv}

\usepackage[utf8]{inputenc} 
\usepackage[T1]{fontenc}    
\usepackage{hyperref}       
\usepackage{url}            
\usepackage{booktabs}       
\usepackage{amsfonts}       
\usepackage{nicefrac}       
\usepackage{microtype}      
\usepackage{lipsum}		
\usepackage{graphicx}
\usepackage{natbib}
\usepackage{doi}
\usepackage{multirow}
\usepackage{kotex}
\usepackage{amsmath}

\begin{document}

\title{MOGAM: A Multimodal Object-oriented Graph Attention Model for Depression Detection}

\author{ \href{https://orcid.org/0009-0005-5153-9668}{\includegraphics[scale=0.06]{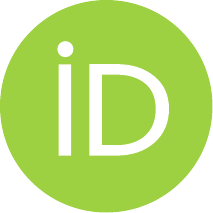}\hspace{1mm}Junyeop Cha} \\
	Department of Applied Artificial Intelligence\\
	Sungkyunkwan University\\
	\texttt{jycha95@g.skku.edu} \\
	\And
	\href{https://orcid.org/0000-0001-5740-897X}{\includegraphics[scale=0.06]{orcid.pdf}\hspace{1mm}Seoyun Kim} \\
	Department of Applied Artificial Intelligence\\
	Sungkyunkwan University\\
	\texttt{seoyun.kim@g.skku.edu} \\
	\AND
	\href{https://orcid.org/0009-0004-4784-2421}{\includegraphics[scale=0.06]{orcid.pdf}\hspace{1mm}Dongjae Kim} \\
	Department of Applied Artificial Intelligence\\
	Sungkyunkwan University\\
	\texttt{bronze.ash@g.skku.edu} \\
        \AND
	\href{https://orcid.org/0000-0002-3177-3538}{\includegraphics[scale=0.06]{orcid.pdf}\hspace{1mm}Eunil Park} \\
	Department of Applied Artificial Intelligence\\
	Sungkyunkwan University\\
	\texttt{eunilpark@skku.edu} \\ 
}



\renewcommand{\headeright}{}
\renewcommand{\undertitle}{}

\hypersetup{
pdftitle={A template for the arxiv style},
pdfsubject={q-bio.NC, q-bio.QM},
pdfauthor={Junyeop Cha, Seoyun Kim, Eunil Park},
pdfkeywords={GNN, Multimodal, Depression},
}

\maketitle

\begin{abstract}
Early detection plays a crucial role in the treatment of depression. Therefore, numerous studies have focused on social media platforms, where individuals express their emotions, aiming to achieve early detection of depression. However, the majority of existing approaches often rely on specific features, leading to limited scalability across different types of social media datasets, such as text, images, or videos. To overcome this limitation, we introduce a Multimodal Object-Oriented Graph Attention Model (MOGAM), which can be applied to diverse types of data, offering a more scalable and versatile solution. Furthermore, to ensure that our model can capture authentic symptoms of depression, we only include vlogs from users with a clinical diagnosis. To leverage the diverse features of vlogs, we adopt a multimodal approach and collect additional metadata such as the title, description, and duration of the vlogs. To effectively aggregate these multimodal features, we employed a cross-attention mechanism. MOGAM achieved an accuracy of 0.871 and an F1-score of 0.888. Moreover, to validate the scalability of MOGAM, we evaluated its performance with a benchmark dataset and achieved comparable results with prior studies (0.61 F1-score). In conclusion, we believe that the proposed model, MOGAM, is an effective solution for detecting depression in social media, offering potential benefits in the early detection and treatment of this mental health condition.
\end{abstract}


\section{Introduction}
Depression is one of the most increasingly severe mental illnesses worldwide. According to the World Health Organization (WHO), about 5\% of adults globally suffer from depression~\cite{who_dep}. Unfortunately, the situation has been exacerbated by the COVID-19 pandemic, which has led to a 25\% global increase in the prevalence of depression. Compounding the problem is the serious issue that not only the majority of individuals do not receive suitable treatment services, but also they are mainly unaware of their conditions~\citep{pramanik2022depression}.

The most representative traditional way of diagnosing depression is a series of face-to-face interviews conducted by psychiatrists, utilizing screening instruments such as CES-D~\citep{radloff1977ces} or PHQ-9~\citep{kroenke2001phq}. However, the COVID-19 pandemic has significantly restricted face-to-face interviews, adding further notable challenges to the diagnostic process~\citep{self2021conducting}. To tackle this problem, a number of researchers have explored the use of social media as an effective man of diagnosing depression~\citep{kim2021machine}.

Social media provides an interactive platform for individuals to share their thoughts, assertions, experiences, and emotions, as well as to connect and communicate with other users. This identified characteristic has allowed social media to be a popular source for analyzing each user's psychological state~\citep{lin2020sensemood,liu2021covid}, particularly in the context of depression detection. Several scholars have utilized various data-driven approaches and technologies, including natural language processing (NLP) and computer vision (CV) techniques, to explore and analyze the user's psychological state from social media~\citep{kim2020deep}.

Representatively, text-based social media platforms like Twitter~\citep{orabi2018deep,cha2022lexicon}, and Reddit~\citep{kim2020deep,ren2021depression} have been extensively explored for this purpose. However, the recent advancements in CV techniques have led to a surge in research emphasized on both image-based and video-oriented social media platforms such as Instagram~\citep{maxim2020predicting} and YouTube~\citep{yoon2022d}. These platforms allow scholars to have new opportunities to examine users' mental states through visual content.

Among various content types in social media platforms, Video blogs (vlogs), are defined as ``\textit{a set of video content on a social media platform for sharing videos}''~\citep{freeman2007youtube}, is one of the most prominent and representative content forms. While vlogs mainly show users' daily lives, there are no restrictions on the content type. This allows users to create vlogs on diverse topics and themes such as hobbies, travel, music, or health. Vloggers (video bloggers) utilize vlogs as one of the tools to present themselves, almost like a personal diary, sharing their thoughts/experiences, and using video technologies to capture and preserve memories~\citep{snelson2015vlogging}. 

In particular, individuals dealing with illnesses often utilize vlogs as a tool to share their conditions and experiences~\citep{huh2014health}. By sharing individuals' states and experiences through vlogs, they can find a supportive community and a platform to express themselves.

In the context of depression detection, prior research has focused on exploring clinical interview videos of individuals with depression~\citep{gratch2014distress}. Facial expressions~\citep{girard2013social}, acoustic signals~\citep{ray2019multi}, and body movements~\citep{joshi2013relative} have been utilized as significant features to capture their psychological state. However, several challenges still exist in this approach. First, datasets based on clinical interviews are costly to obtain, resulting in a limited number of samples available for analysis. Additionally, models trained on such datasets may not be applicable to real-world scenarios or datasets. For instance, if a person's face or body is not detected or obscured by other objects, it becomes difficult to extract significant features from the videos.

To address these challenges, we propose a novel approach called MOGAM, a multimodal object-oriented graph attention model, for depression detection with vlogs regardless of the specific conditions or constraints. We collected depression and non-depression videos from YouTube vlogs using relevant hashtags. It allows us to create high-risk depression and depression datasets consisting of vlogs uploaded before and after users were clinically diagnosed with depression. That is, our research question (RQ) is presented as follows: 

\begin{itemize}
    \item \textbf{RQ}: Can we accurately detect depression and high-risk depression in vlogs using our proposed method?
\end{itemize}

Considering RQ, in addition to employing specific features like facial expressions, we employed a unique approach. We extract objects presented in each vlog (e.g. person, cup, bed), and create an object network by computing the co-occurrence count between pairs of objects. This forms the basic framework of MOGAM, which leverages the object co-occurrence network to extract features from vlogs.

Our model utilizes a graph neural network (GNN), which is designed to learn representations of nodes or graphs based on their structural relationships. In this case, GNN operates on the object co-occurrence network, allowing it to capture the interactions among different objects within a vlog. Moreover, we extract additional features such as visual information or metadata from vlogs. These features are aggregated with the object-based graph features using a cross-attention mechanism. It enables the model to learn relevant features from different modalities, integrating them to improve the overall representation and depression detection in vlogs.

We also introduce a dataset including 4767 vlogs, which encompass both depression and high-risk potential groups. These vlogs serve as the foundation of our experiments. We demonstrate that our proposed approach is robust with no dependence on specific features in vlogs such as human faces, poses, or eye movements. Thus, this flexibility allows to be applied to other video datasets, expanding its utility and potential for broader use cases. We publicly release our collected datasets and the MOGAM architecture.

\section{Related Work}
\label{sec:headings}
\subsection{High-risk mental disorders detection}
Proactively identifying individuals at risk of mental disorders is crucial because early diagnosis is one of the most important issues in effective treatments ~\citep{conus2014public}. To achieve such a goal, data science and machine learning came to play an essential role in identifying potential symptoms and risk factors, as they are strongly correlated to mental disorders~\citep{thieme2020machine}. For example, \citet{hao2013predicting} aimed to identify users' mental health status through social media, leveraging machine learning to detect at-risk individuals. Similarly, \citet{wang2017detecting} built a predictive model for eating disorders using Twitter data, analyzing social status, behavioral patterns, and psychometric properties of individuals in the disorder group and non-disorder group.

Numerous studies have also investigated early detection and risk prediction for several notable mental disorders since the impact of depression on individuals and society is growing significantly, and early detection and diagnosis remain crucial for effective treatment~\citep{halfin2007depression}. \citet{xu2019individualized} and \citet{king2008development} have conducted extensive research on detecting high-risk groups for depression by analyzing symptoms and patterns associated with the disorder. However, they commonly face challenges such as time-consuming experiments, limited sample sizes, and relatively high costs. Several researchers have turned to social media platforms to overcome these limitations as a valuable potential source for detecting depression. This approach takes advantage of these platforms' vast amount of user-generated content, providing critical opportunities for more scalable and cost-effective detection methods~\citep{cha2022lexicon,kim2020deep,de2013predicting}.

\subsection{Depression Detection in Social Media}
Utilizing social media datasets in terms of detecting depression provides some distinct advantages, such as enabling the researchers to make use of large datasets and to apply more sophisticated data-driven approaches. These features greatly help the understanding of an individual’s mental health, which leads to the vast usage of social media in depression detection research~\citep {de2013role,de2015social,balani2015detecting}. For example, \citet{al2019depression} and \citet{lin2020sensemood} employed Facebook and Twitter datasets to distinguish depression users using machine learning and deep learning approaches, respectively. Some other scholars used Reddit, the topic-oriented social media services, to apply various approaches such as linear SVM with bag-of-n-gram features~\citep{pirina2018identifying} or MLP on LIWC, LDA, and bi-gram features~\citep{tadesse2019detection}. Most of these studies highlighted the data-driven approaches for detecting depression in social media, indicating the potentiality of social media as a valuable resource.

Video is one of the primarily modality for detecting depression, offering diverse features such as visual cues. Especially, user-oriented features presented in video content, including facial appearance and pose, are commonly adopted for depression detection. For example, \citet{guo2022automatic} fed 2D landmarks and head pose features from the DAIC-WOZ dataset to the CNN-based model, achieving an accuracy of 0.857. \citet{wang2018facial} focused on Chinese individuals and employed clinical video samples to detect depression, achieving an accuracy of 0.789 using SVM with facial expression and eye movement features. Another mainstream of utilizing video data on depression detection is the multimodal approach, which incorporates various features such as visual, audio, and metadata information from given videos. \citet{yoon2022d} and \citet{chen2021sequential} also applied a multimodal fusion model on facial visual and acoustic features to identify people’s depression.
While prior studies primarily rely on human-oriented features like facial expressions, eye movement, or pose, one of the challenges is their limited applicability to videos without human presence. This poses a limitation for real-world applications since many vlogs may not necessarily involve people in their content. In contrast, the proposed object-based graph method is not restricted to human-appearing videos and can be applied to any type of video. To the best of our knowledge, your study is the first to examine depression detection in vlogs using an object-based Graph Neural Network (GNN) approach, which provides a novel perspective for addressing this important problem.

\section{MOGAM: Multimodal Object-oriented Graph Attention Model}
In this section, we introduce the construction process of our vlog dataset and the classification method. Specifically, we cover (1) the procedure for collecting and preprocessing YouTube vlogs, and (2) the introduction of our multimodal object-oriented graph attention model: MOGAM.

\subsection{Data Collection and Preprocessing}
To collect vlogs, we utilized the YouTube API (Application Programming Interface). We searched vlogs using two specific hashtags: ``\#우울증브이로그'' (depression vlog) and ``\#일상브이로그'' (daily vlog). All daily vlogs resulting from the search using the ``\#일상브이로그'' hashtag were collected without any additional filtering procedures. However, for the depression vlogs indicated by the ``\#우울증브이로그'' hashtag, we manually inspected the results because the hashtag alone could not guarantee the vlogs' relevance to a medical depression diagnosis. Thus, the following steps were conducted.

\begin{figure*}[t]
    \begin{center}
    \includegraphics[width=0.75\linewidth]{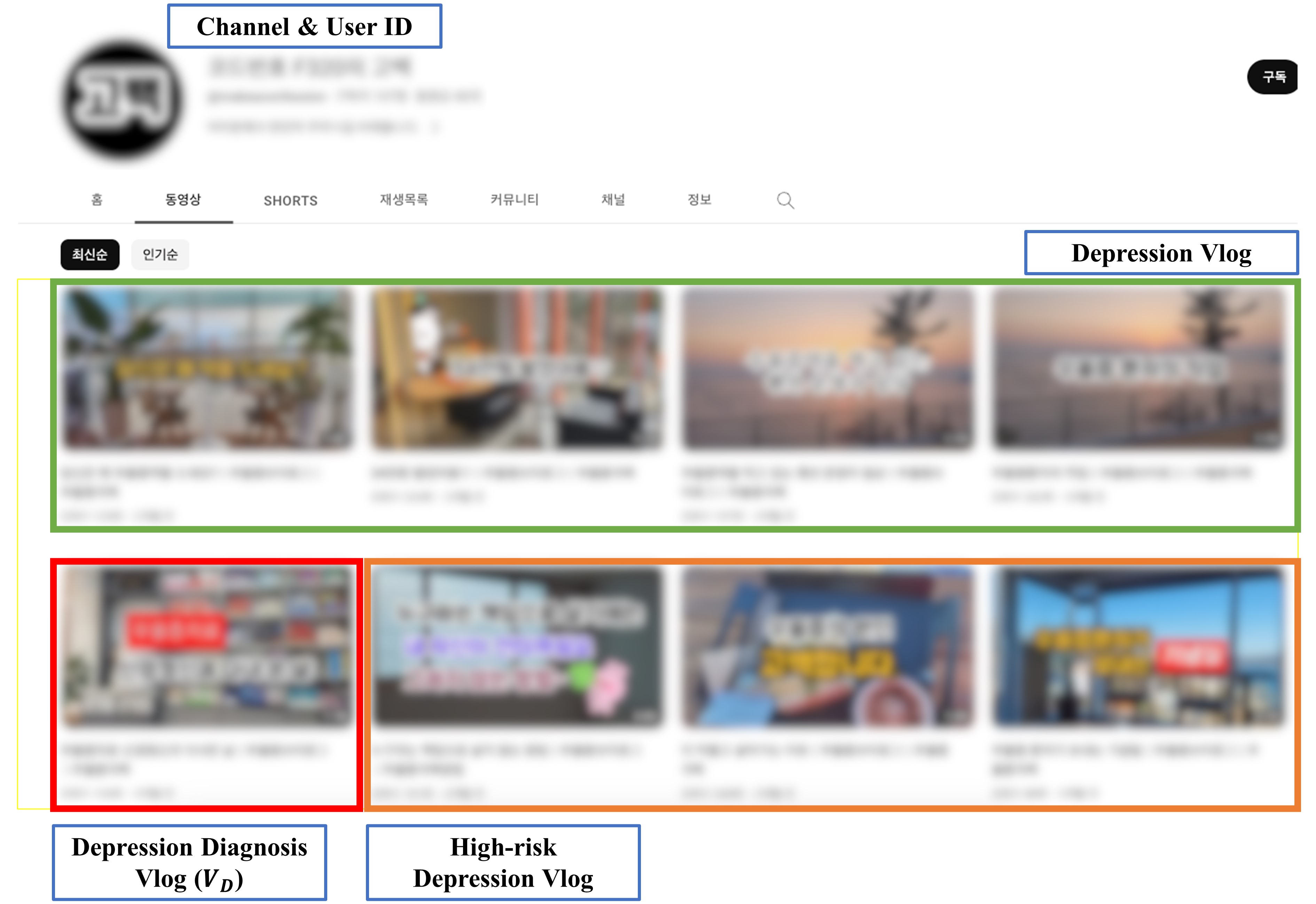}
    \caption{Examples of the collected datasets.}
    \label{fig:example}
    \end{center}
\end{figure*} 

\begin{enumerate}
    \item All vlogs uploaded by users, who wrote the ``\#우울증브이로그'' hashtag, were collected.
    \item Two separate researchers carefully reviewed the content of each user's vlogs to identify the first collected vlog, which indicated a diagnosis vlog ($V_{D}$). 
    \item Based on the upload time of each depression diagnosis vlog $V_{D}^{(t)}$, we divided the user's vlog list into two groups: high-risk depression vlogs, which were uploaded before $V_{D}$, and depression vlogs, which were uploaded after $V_{D}$. Figure~\ref{fig:example} provides the representative cases demonstrating the vlog collection and division procedures.
\end{enumerate}

Based on these procedures, the resulting dataset is organized by 1888 daily, 2237 depression, and 642 high-risk depression vlogs. The dataset consists of three groups: daily, depression, and high-risk depression. The daily group includes vlogs uploaded by non-depressed individuals unrelated to depression. The depression group consists of vlogs from individuals clinically diagnosed with depression. The high-risk depression group includes vlogs created by individuals who have not received a clinical diagnosis of depression but may exhibit symptoms associated with depression. The average duration times of these vlogs are 903.39, 416.03, and 515.74 seconds, respectively (Table~\ref{tab:vlog_stat}). 

\begin{table}[htb]
    \centering 
    \begin{tabular}{l | r r}
        \hline
                                &  \# of Vlogs &   Avg. duration    \\
        \hline 
        Daily                   &   1888    &   903.39s          \\
        Depression              &   2237    &   416.03s          \\
        High-risk depression    &   642     &   515.74s          \\
        \hline
    \end{tabular}
    \centering
    \caption{Descriptive statistics of the collected dataset.}~\label{tab:vlog_stat}
\end{table}

To analyze vlogs for depression detection, we utilized both image frames and metadata. Initially, we split each vlog into frames at a rate of single frame per second (FPS). It resulted in transforming the vlogs into a collection of individual images. In addition to the image frames, we gathered relevant metadata from the vlogs, including the title, description and duration. It is worth noting that while providing a description for a vlog is not mandatory on Youtube, we encountered cases where no description was available. In such instances, we replaced the missing description with empty string as dummy input for ensuring consistency in the metadata collection procedures.

\subsection{Depression Detection on the vlog}
The proposed model, MOGAM, is organized by three key components for depression detection in vlogs as presented as follows: 1) our object-oriented graph neural network, 2) the extraction of additional visual and metadata features, and 3) the aggregation of multimodal features to detect depression in vlogs. Through the integration of the object-oriented GNN, visual, and metadata features, our proposed model provides a comprehensive approach for depression detection in vlogs, leveraging both the inherent patterns within the vlogs and the multimodal information available.

\subsubsection{Object-oriented Graph Neural Network}

\begin{figure*}[t]
    \centering
    \includegraphics[width=0.95\linewidth]{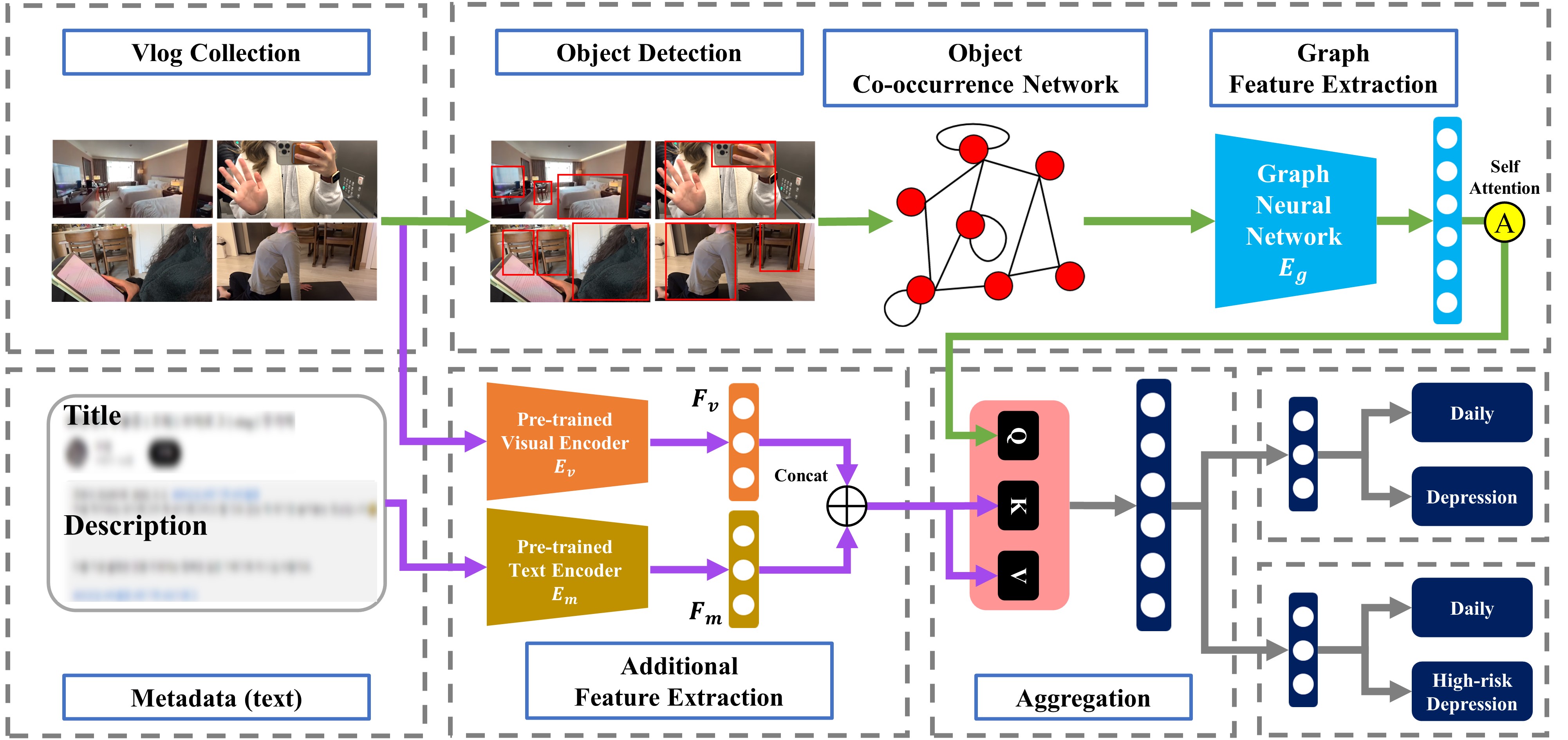}
    \caption{The overview of MOGAM; It is organized by three parts: data collection, feature extraction, and detection.}
    \label{fig:model}
\end{figure*} 

To construct the vlog object adjacency matrix, we conducted the following steps:

\begin{enumerate}
    \item \textbf{Detection of objects in frames}: We employed YOLOv5~\citep{glenn_jocher_2020_4154370}, one of the widely-employed open source object detection models, to detect objects in each frame of the vlogs. 
    \item \textbf{Defining nodes and edges}: We defined the detected the objects and co-occurrence count of each object pair in a frame as nodes and edges, respectively. If multiple objects of the same class were found in a frame (e.g. two different cup), we considered them as separate nodes. 
    \item \textbf{Construction of adjacency matrix}: To build the adjacency matrix, we utilized the co-occurrence counts of the object pairs.
    \item \textbf{Normalization}: Because the duration of vlogs may vary, the adjacency matrix was normalized by dividing each entry by the total number of frames. It ensures that the graph structure remains consistent irrespective of the length of vlogs.
\end{enumerate}

Based on these steps, we got a weighted undirected adjacency matrix which represents the object co-occurrence in the vlog. This matrix can capture the relationships among different objects, and serve as a basis for comparing the graph structures of other vlogs. The construction of the vlog object adjacency matrix is on of the fundamental steps in our approach to investigate the vlog depression detection. That is, the following equation is presented for building the proposed vlog object adjacency matrix ($A\in \mathbb{R}^{T \times T}$):
\begin{align}
    & a^{(v)}_{ij} = \sum_{n=1}^{N^{(v)}}{\sum_{o_{i},o_{j} \in T}{f^{(v)}_{n}(o_i, o_j)}} \\
    & \Tilde{a}^{(v)}_{ij} = \frac{1}{N^{(v)}}a^{(v)}_{ij}
\label{adj_mat_equation}
\end{align}

where $N^{(v)}, f^{(v)}_{n}(\cdot)$ denote the number of frames in the $v$th vlog, the function $f^{(v)}_{n}(\cdot)$ returns the co-occurrence count of two objects ($o_i, o_j$) in $n$th frame of $v$th vlog, respectively. Next, we fed $A$ into three off-the-shelf GNN models: GCN~\citep{kipf2016semi}, GraphSAGE~\citep{hamilton2017inductive}, and GAT~\citep{velivckovic2017graph}. We compare the results of our proposed framework, MOGAM with the existing GNN models. The detailed description of the procedures are presented as follows:
\begin{align}
    \tag{GCN}
    &h_{t}^{(l+1)} = \sigma(W_l \sum_{u \in N(t)} \frac{h_{u}^{(l)}}{\left\vert N(t) \right\vert} + B_lh_{t}^{(l)}) \\
    \tag{GraphSAGE}
    &h_{t}^{(l+1)} = \sigma(\left[W_l \cdot AGG\left(\{h_{u}^{(l)} \vert u \in N(t)\}\right), B_lh_{t}^{(l)}\right])\\
    \tag{GAT}
    &h_{t}^{(l+1)} = \sigma(W_l \sum_{u \in N(t)} \alpha_{ut}h_{t}^{(l)})
\end{align}\label{gnn_equation}

where $t, u$ are nodes and $\sigma, h_{t}^{(l)}, N(t), W_l, B_l, \alpha, AGG$ denote the nonlinear function, node $t$'s feature vectors at $l$th layer, neighbors of node $t$, information of neighbor's and myself in the previous layer, attention weight, and aggregation function, respectively. Next, we fed $A$ and node feature matrix made up of one-hot vectors into a number of GNN layers. Then, to obtain graph-level representation, we applied a global mean pooling, which involves averaging the node features across the node dimension. This pooling operation makes sure that the output representation can effectively capture the overall vlog information.

\subsubsection{Visual Feature}
To extract visual features, we utilized a pre-trained ResNet~\citep{he2016deep}, which is commonly employed for transfer learning and image feature extraction. We extracted feature vectors with a dimension of 1,000 and reduced them to the same size as the object-oriented graph feature using a fully connected layer. To ensure that the extract features are independent of the vlog length, we averaged all feature vectors.

\subsubsection{Metadata Feature}
For metadata feature extraction, we chose the title, description and duration from various metadata, which  is available on Youtube, one of the globally used social media platforms. These features are selected by the eusers during the vlog uploading procedures. To prepare the metadata, we removed unnecessary textual information such as email addresses, URLs, and non-Korean text via data pre-processing steps. Consequently, we utilized pre-trained KoBERT~\citep{kobert}.

Note that if other multilingual language models such as M-BERT~\citep{pires-etal-2019-multilingual} or XLMs~\citep{lample2019cross} were adopted, MOGAM would be applicable to various languages. The feature vectors obtained from the title, description, and duration of the vlog were concatenated into a single feature vector.

\subsubsection{Aggregation \& Detection}
Each encoder ($E_g$, $E_v$, and $E_m$) generates features $F_g \in \mathbb{R}^{d}$ and $F_v,\ F_m \in \mathbb{R}^{d/2}$ where $d$ denotes the hidden dimension (Figure~\ref{fig:model}). Then, we concatenated visual and metadata features to build additional integrated features.
\begin{align}
    F_a = F_v \oplus F_m
\end{align} 

To enhance the model's ability to capture the inherent patterns of vlogs, which may not be captured by the baseline models, we incorporated a cross-attention mechanism within a transformer architecture~\citep{vaswani2017attention}. In this mechanism, we utilized $F_g$ as the query (Q) and $F_a$ as the key (K) and value (V) for the cross-attention module.
\begin{align}
    &ATEENTION(Q, K, V) = softmax(\frac{QK^T}{\sqrt{d_k}}V) \\
    &F = Attention(F_{g}, F_{a}, F_{a})
\end{align}

To prevent an over-fitting tendency, we applied a dropout regularization. We employed several fully connected layers to reduce the dimensionality of the integrated multimodal features, $F$, for vlog classification. The reduced features were then passed through a sigmoid function. The output represents the logit of the corresponding label.

\section{Experiments}
We conducted three conditional experiments as follows:

\begin{itemize}
    \item Daily versus Depression
    \item Daily versus High-risk Depression
    \item Scalability Evaluation
\end{itemize}

In addition to conducting a series of experiments using the multimodal approach, we implemented the experiments not only using the unimodal ($F_g$) approach, but also employing multimodal information for examining the effects of features on the model performance.



All experiments were conducted using a single NVIDIA RTX A6000 48GB GPU and Python 3.7. We used the Yolo v5 model, which was pre-trained on the COCO dataset~\citep{lin2014microsoft}, which consists of 80 objects. We split both datasets into the train, validation, and test sets in an 8:1:1 proportion (five times). The exact number of each set is presented in the appendix. We used the Adam optimizer~\citep{kingma2014adam} and set batch size, epochs, and hidden dimensions to 32, 500, and 1024, respectively.

\begin{table*}[htbp]
    \centering
    \scalebox{0.95}{
    \begin{tabular}{p{2.75cm}|c|c|rrrr}
        \hline
        & Model & Label &    Accuracy    &  Precision   &   Recall   &   F1-Score    \\
        \hline
        \multirow{12}{2.75cm}{Daily and Depression} & \multirow{2}{*}{GCN} & Daily  & \multirow{2}{*}{0.802} &  0.800 & 0.784    &   0.792 \\
        & & Depression  &  & 0.803   &   0.818 &   0.810 \\
        \cline{2-7}
        & \multirow{2}{*}{GraphSAGE} & Daily   &  \multirow{2}{*}{0.794}  &  0.779 & 0.799    &   0.789 \\
        & & Depression  &   & 0.809   &   0.790 &   0.799 \\
        \cline{2-7}
        & \multirow{2}{*}{GAT}  & Daily   &  \multirow{2}{*}{0.826}  &  0.798 & 0.854    &   0.825 \\
        & & Depression  &    & 0.855   &   0.799 &   0.826 \\
        \cline{2-7}
        & \multirow{2}{*}{MOGAM with GCN} &  Daily   &  \multirow{2}{*}{0.839}  &  0.787 & 0.851    &   0.818\\
        & & Depression  &   & 0.883   &   0.831 &   0.857 \\
        \cline{2-7}
        & \multirow{2}{*}{MOGAM with GraphSAGE} & Daily   &  \multirow{2}{*}{0.864}  &  0.869 & 0.799    &   0.832 \\
        & & Depression  &    & 0.861   &   \textbf{0.911} &   0.885 \\
        \cline{2-7}
        & \multirow{2}{*}{MOGAM with GAT} & Daily   &  \multirow{2}{*}{\textbf{0.871}}  &  0.850 & 0.845    &   0.847 \\
        & & Depression  &    & \textbf{0.887}   &   0.890 &   \textbf{0.888} \\
        \hline
        \multirow{12}{2.75cm}{Daily and High-risk Depression} & \multirow{2}{*}{GCN} 
        & Daily   &  \multirow{2}{*}{0.811}  &  0.832 & 0.930    &   0.878 \\
        & & High-risk potential  &   & 0.717   &   0.485 &   0.579 \\
        \cline{2-7}
        & \multirow{2}{*}{GraphSAGE}  
        & Daily   &   \multirow{2}{*}{0.771}  &  0.814 & 0.893   &   0.851 \\
        & & High-risk potential  &    & 0.600   &   0.441 &   0.509 \\
        \cline{2-7}
        & \multirow{2}{*}{GAT}  
        & Daily   & \multirow{2}{*}{0.823}   &  0.844 & 0.930    &   0.885 \\
        & & High-risk potential  &    & 0.735   &   0.529 &   0.615 \\        
        \cline{2-7}
        & \multirow{2}{*}{MOGAM with GCN}
        & Daily   &  \multirow{2}{*}{0.996}  &  0.995 & 1.000    &   0.997 \\
        & & High-risk potential  &    & 1.000   &   0.986 &   0.993 \\
        \cline{2-7}
        & \multirow{2}{*}{MOGAM with GraphSAGE}
        & Daily   &  \multirow{2}{*}{0.996}  &  0.995 & 1.000    &   0.997 \\
        & & High-risk potential  &    & 1.000  &   0.986 &   0.993\ \\
        \cline{2-7}
        & \multirow{2}{*}{MOGAM with GAT}
        & Daily                &  \multirow{2}{*}{\textbf{0.996}}  &  0.995 & \textbf{1.000}     &   \textbf{0.997} \\
        & & High-risk potential  &    & \textbf{1.000}      & 0.986 &   0.993 \\       
        \hline
    \end{tabular}}
    \centering
    \caption{Performance comparisons between baselines and proposed models for our datasets.}~\label{tab:depression_result}
\end{table*}

\subsection{Daily \& Depression}

The performance of the proposed model on both daily and depression vlogs is summarized in Table~\ref{tab:depression_result}. Except for GraphSAGE, all the proposed models exhibited significant accuracy and F1-score, providing evidence for the effectiveness of GNN-based models in classifying depression vlogs. Notably, MOGAM with GAT achieved the highest F1-score among the baselines, indicating that the cross-attention mechanism is a suitable approach for learning representations based on the relationships between objects. This finding underscores the importance of capturing inter-object relationships in effectively detecting depression patterns. Consistent with the findings of prior work~\citep{yoon2022d}, the models incorporating multimodal features demonstrated superior performance compared to the baselines. This finding suggests that the integration of object-based graph features with visual and metadata features creates a robust framework for effectively identifying depression in vlogs.

\subsection{Daily \& High-risk Potential}

To learn the symptoms and patterns of the high-risk depression state prior to clinical diagnosis, we trained our models using the daily and high-risk potential datasets. The performance comparisons on daily and high-risk depression vlogs are also presented in Table~\ref{tab:depression_result}. The data structure and classification model architecture for the daily vlogs are identical to those described in the previous section. We observed that incorporating multimodal features enables the model to capture the distinctions between daily and high-risk depression vlogs. Importantly, with the implementation of MOGAM, all models achieved accurate discrimination between daily and high-risk depression vlogs (F1-score of 0.997).

\subsection{Multiclass Classification}
Finally, a multiclass classification experiment was conducted within the same experimental environment as the previous sections, aiming to simultaneously learn and distinguish symptoms and patterns across all three states. The models were trained on datasets representing daily, depression, and high-risk potential states, and the comparison of results is presented in Table~\ref{tab:multiclass}. In contrast to the binary classification, the GraphSAGE model combined with MOGAM exhibited the best F1-score of 0.815 and 0.750 for daily and high-risk potential states, respectively while the GAT model with MOGAM presented superior performance in the depression group with an F1-score of 0.783. However, the GCN, GraphSAGE, and GAT models without MOGAM exhibited limited ability to distinguish high-risk potential states, mostly misclassifying them as daily and depression (F1-score: 0.061, 0.083, 0.182). In contrast, the proposed MOGAM ensured comprehensive classification of high-risk potential states across all models (F1-score: 0.713, 0.750, 0.736). Consequently, we observed that MOGAM is suited for capturing and classifying image-specific features of each state simultaneously, even in an imbalanced data environment.

\begin{table*}[ht]
    \centering 
    \begin{tabular}{p{4cm} | l | r r r r}
    
    \hline
    Model             & Label               & Accuracy               & Precision & Recall & F1-Score \\
    \hline
    \multirow{3}{4cm}{MOGAM with GCN} & Daily    & \multirow{3}{*}{\textbf{0.790}} & \textbf{0.832}     & 0.806  & 0.819 \\
                         & Depression          &                        & \textbf{0.724}     & 0.849  & 0.782 \\
                         & High-risk potential &                        & 0.973     & 0.563  & 0.713 \\
    \hline                           
    \multirow{3}{4cm}{MOGAM with GraphSAGE} & Daily    & \multirow{3}{*}{0.788} & 0.789     & \textbf{0.843}  & \textbf{0.815} \\
                                          & Depression &                      & 0.749     & 0.792  & 0.770 \\
                                          & High-risk potential &             & \textbf{0.975}     & \textbf{0.609}  & \textbf{0.750} \\                           
    \hline
    \multirow{3}{4cm}{MOGAM with GAT} & Daily    & \multirow{3}{*}{0.785} & \textbf{0.832}     & 0.775  & 0.802 \\
                         & Depression          &                        & 0.723     & \textbf{0.854}  & \textbf{0.783} \\
                         & High-risk potential &                        & 0.929    & \textbf{0.609}  & 0.736 \\
    \hline
    \end{tabular}%
    \caption{Results of the multiclass classification. The train dataset consists of 1,507 daily, 1,567 depression, and 502 high-risk potential vlogs and the test dataset consists of 191, 192, and 64. The results of baselines are presented in appendix.}~\label{tab:multiclass}
\end{table*}

\subsection{Scalability Evaluation}
\begin{figure}[ht]

    \centering
    \includegraphics[width=.4\textwidth]{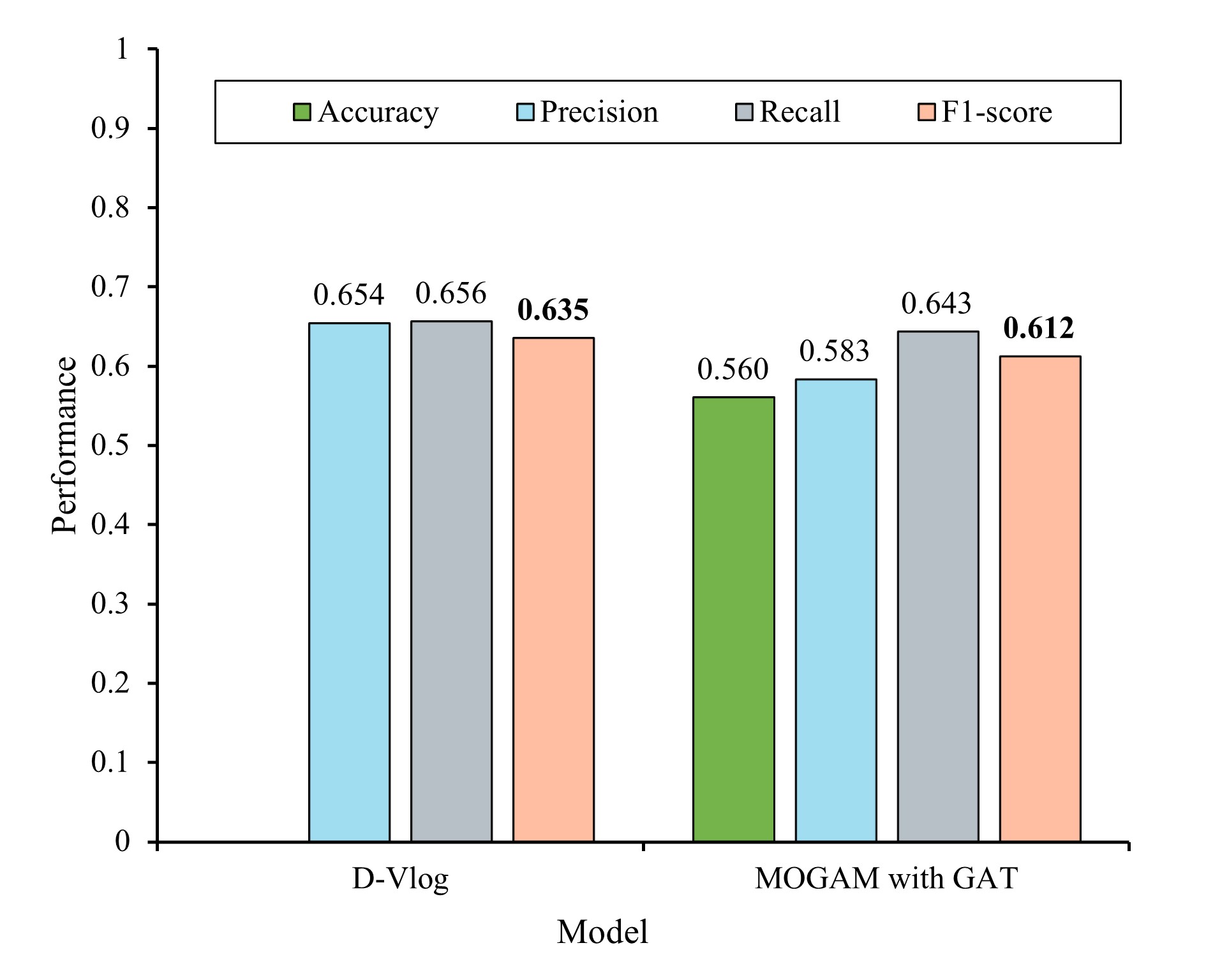}
    \caption{Performance evaluation of prior multimodal fusion models and MOGAM with GAT on D-Vlog dataset}
    \label{fig:dvlog}
\end{figure}

To address the scalability of the proposed methods, we conducted experiments on D-Vlog~\citep{yoon2022d}, which is used in several prior research for depression detection in vlogs. It consists of 555 depression and 406 non-depression English vlogs. To conduct a fair evaluation, we followed same procedures to extract object-based graph, visual and metadata features from D-Vlog. Subsequently, we applied the models trained on our vlog dataset, including both baseline and multimodal features, to D-Vlog dataset. Among these models, the GAT-based model trained with multimodal features achieved the highest F1-score of 0.612, which is comparable to the reported results in prior research (0.635). Figure~\ref{fig:dvlog} shows the comparison results.


\subsection{Object Distribution}
To investigate gaps in the frequency of object appearance across different state vlogs, we conducted one-way analyses of variance (ANOVA) and post hoc tests. Employing an ANOVA involving 80 objects, we identified 47 objects exhibiting statistically significant differences in their mean values among the three groups. Subsequently, the results of post hoc tests present that the object ``fork" exhibited significant differences in all groups, while only eight objects (knife, cake, handbag, sandwich, apple, wine glass, banana, and vase) exhibited differences between the daily and non-daily group (depression and high-risk potential). The objects associated with food demonstrated notably low p-values, indicating a strong association between food-related objects and the user's states in vlogs. Moreover, the frequency of object appearance is higher in the daily group compared to the non-daily group. The mean and standard deviation (SD) normalized appearance counts for the five objects with the lowest p-value are presented in Table~\ref{tab:object}. The entire results of analysis and the post hoc test are shown in the appendix.
 
\begin{table}[ht]
    \centering 
    \begin{tabular}{l | c | r r}
    \hline
    Object & Label & Mean (SD) \\
    \hline    
    \multirow{3}{*}{Knife} & Daily & 0.284 (0.591)\\ \cline{2-3}
          & Depression & 0.068 (0.291)\\ \cline{2-3}
          & High-risk potential  & 0.079 (0.332) \\
    \hline
    \multirow{3}{*}{Fork}& Daily & 0.240 (0.517) \\ \cline{2-3}
          & Depression & 0.071 (0.161) \\ \cline{2-3}
          & High-risk potential  & 0.374 (0.631) \\
    \hline
    \multirow{3}{*}{Cake}& Daily & 0.340 (1.286) \\ \cline{2-3}
          & Depression & 0.078 (0.530) \\ \cline{2-3}
          & High-risk potential  & 0.093 (0.451) \\
    \hline
    \multirow{3}{*}{Handbag}& Daily & 0.527 (1.316)\\ \cline{2-3}
          & Depression & 0.211 (1.670) \\ \cline{2-3}
          & High-risk potential  & 0.298 (1.549) \\
    \hline
    \multirow{3}{*}{Sandwich}& Daily & 0.130 (0.562) \\ \cline{2-3}
          & Depression & 0.040 (0.314) \\ \cline{2-3}
          & High-risk potential  & 0.056 (0.310) \\   
    \hline    
    \end{tabular}
    \centering
    \caption{Descriptive information of top-object appearances.}~\label{tab:object}
\end{table}

\section{Discussion}
In this study, we proposed MOGAM, a new multimodal object-oriented graph attention model for depression detection on YouTube vlogs. To facilitate our approach, we collected a vlog dataset including vlogs of daily, depression, and high-risk potential depression groups. Then, we extracted three key multimodal features (graph, visual, and metadata) from the vlogs and combined them using a cross-attention mechanism. By leveraging these integrated multimodal features, our proposed model aims to determine whether the vlog uploader is experiencing depression. The results on the dataset demonstrate that our model effectively captures and detects depression symptoms and patterns in the vlogs by leveraging the power of multimodal features.

Based on our findings, we can draw several implications. Many researchers utilize social media platforms for depression detection, leading to the development of various depression-related datasets~\citep{gratch2014distress,yoon2022d}. However, there is often a trade-off between dataset size and reliability. Building a dataset with clinical diagnoses can be prohibitively expensive, making it challenging to create large-scale datasets of this nature. In contrast, our vlog dataset comprises vlogs from individuals who have undergone clinical diagnosis, providing valuable insights into real depression patterns and symptoms. Moreover, our study validates the effectiveness of the object-oriented graph encoder and multimodal features for depression detection in YouTube videos. The scalability experiment results suggest that our proposed MOGAM can be applied to other datasets, including those in different languages and environments.

Although we present several implications, notable concerns remain. First, the extraction of object-oriented graph features depends on the performance of the employed object detection model (e.g., Yolo v5). If the object detection model poorly performs, the constructed object co-occurrence network may not accurately represent the entire vlog, that can lead to the lower performance level of MOGAM. Second, because additional features we employed can depend on off-the-shelf models, we extracted them using pre-trained encoders (ResNet and KoBERT), which were trained on other datasets. Therefore, the quality and performance of these off-the-shelf models can impact the effectiveness of the additional features in our framework.

In future research, we aim to extend our method. We could apply state-of-the-art (SOTA) object detection methods. The SOTA object detection model can take advantage of a large number of objects and accurate detection performance, letting the model to capture complex relationships among objects. In addition, by utilizing encoders such as Transformer, we can effectively encode multimodal features and generate integrated representations of vlogs, leading to the improved performance. Furthermore, since depression is not the only mental disorder, we could be valuable to collect vlogs related to other mental disorders (e.g. anxiety, BPD, bipolar disorder) and apply our models to them. 

We believe that our dataset and method as valuable tools for gaining insights into potential depression symptoms in vlogs. Consequently, wee hope that our model can assist individuals who are suffering from depression but may be unaware of their symptoms, as well as those who are not receiving adequate treatment. By effectively detecting inherent depression patterns in their vlogs, our model has the potential to offer support and guidance to those in need, ultimately leading to improved mental health outcomes.

\bibliographystyle{unsrtnat}
\bibliography{references}  

\begin{thebibliography}{49}
\providecommand{\natexlab}[1]{#1}
\providecommand{\url}[1]{\texttt{#1}}
\expandafter\ifx\csname urlstyle\endcsname\relax
  \providecommand{\doi}[1]{doi: #1}\else
  \providecommand{\doi}{doi: \begingroup \urlstyle{rm}\Url}\fi

\bibitem[{WHO}(2023)]{who_dep}
{WHO}.
\newblock Depressive disorder (depression).
\newblock \url{https://www.who.int/news-room/fact-sheets/detail/depression}, 2023.
\newblock Accessed: 2023-06-08.

\bibitem[Pramanik et~al.(2022)Pramanik, Bijoy, and Rahman]{pramanik2022depression}
Anik Pramanik, Md~Hasan~Imam Bijoy, and Md~Sadekur Rahman.
\newblock Depression-level prediction during covid-19 pandemic among the people of bangladesh using ensemble technique: Mirf stacking and mirf voting.
\newblock In \emph{Proc. of IC4IR '21}, pages 71--87. Springer, 2022.

\bibitem[Radloff(1977)]{radloff1977ces}
Lenore~Sawyer Radloff.
\newblock The ces-d scale: A self-report depression scale for research in the general population.
\newblock \emph{Applied psychological measurement}, 1\penalty0 (3):\penalty0 385--401, 1977.

\bibitem[Kroenke et~al.(2001)Kroenke, Spitzer, and Williams]{kroenke2001phq}
Kurt Kroenke, Robert~L Spitzer, and Janet~BW Williams.
\newblock The phq-9: validity of a brief depression severity measure.
\newblock \emph{Journal of general internal medicine}, 16\penalty0 (9):\penalty0 606--613, 2001.

\bibitem[Self et~al.(2021)]{self2021conducting}
Becky Self et~al.
\newblock Conducting interviews during the covid-19 pandemic and beyond.
\newblock In \emph{Forum Qualitative Sozialforschung/Forum: Qualitative Social Research}, volume~22. DEU, 2021.

\bibitem[Kim et~al.(2021)Kim, Lee, Park, et~al.]{kim2021machine}
Jina Kim, Daeun Lee, Eunil Park, et~al.
\newblock Machine learning for mental health in social media: bibliometric study.
\newblock \emph{Journal of Medical Internet Research}, 23\penalty0 (3):\penalty0 e24870, 2021.

\bibitem[Lin et~al.(2020)Lin, Hu, Su, Li, Mei, Zhou, and Leung]{lin2020sensemood}
Chenhao Lin, Pengwei Hu, Hui Su, Shaochun Li, Jing Mei, Jie Zhou, and Henry Leung.
\newblock Sensemood: depression detection on social media.
\newblock In \emph{Proc. of ICMR '20}, pages 407--411, 2020.

\bibitem[Liu et~al.(2021)Liu, Liu, Yoganathan, and Osburg]{liu2021covid}
Hongfei Liu, Wentong Liu, Vignesh Yoganathan, and Victoria-Sophie Osburg.
\newblock Covid-19 information overload and generation z's social media discontinuance intention during the pandemic lockdown.
\newblock \emph{Technological Forecasting and Social Change}, 166:\penalty0 120600, 2021.

\bibitem[Kim et~al.(2020)Kim, Lee, Park, and Han]{kim2020deep}
Jina Kim, Jieon Lee, Eunil Park, and Jinyoung Han.
\newblock A deep learning model for detecting mental illness from user content on social media.
\newblock \emph{Scientific reports}, 10\penalty0 (1):\penalty0 1--6, 2020.

\bibitem[Orabi et~al.(2018)Orabi, Buddhitha, Orabi, and Inkpen]{orabi2018deep}
Ahmed~Husseini Orabi, Prasadith Buddhitha, Mahmoud~Husseini Orabi, and Diana Inkpen.
\newblock Deep learning for depression detection of twitter users.
\newblock In \emph{Proc. of CLPsych '18}, pages 88--97, 2018.

\bibitem[Cha et~al.(2022)Cha, Kim, and Park]{cha2022lexicon}
Junyeop Cha, Seoyun Kim, and Eunil Park.
\newblock A lexicon-based approach to examine depression detection in social media: the case of twitter and university community.
\newblock \emph{Humanities and Social Sciences Communications}, 9\penalty0 (1):\penalty0 1--10, 2022.

\bibitem[Ren et~al.(2021)Ren, Lin, Xu, Zhang, Yang, and Sun]{ren2021depression}
Lu~Ren, Hongfei Lin, Bo~Xu, Shaowu Zhang, Liang Yang, and Shichang Sun.
\newblock Depression detection on reddit with an emotion-based attention network: algorithm development and validation.
\newblock \emph{JMIR Medical Informatics}, 9\penalty0 (7):\penalty0 e28754, 2021.

\bibitem[Maxim et~al.(2020)Maxim, Smirnov, and Ignatiev]{maxim2020predicting}
S~Maxim, I~Smirnov, and N~Ignatiev.
\newblock Predicting depression with social media images.
\newblock In \emph{Proc. of ICPRAM '20}, pages 235--240, 2020.

\bibitem[Yoon et~al.(2022)Yoon, Kang, Kim, and Han]{yoon2022d}
Jeewoo Yoon, Chaewon Kang, Seungbae Kim, and Jinyoung Han.
\newblock D-vlog: Multimodal vlog dataset for depression detection.
\newblock In \emph{Proc. of AAAI '22}, volume~36, pages 12226--12234, 2022.

\bibitem[Freeman and Chapman(2007)]{freeman2007youtube}
Becky Freeman and Simon Chapman.
\newblock Is “youtube” telling or selling you something? tobacco content on the youtube video-sharing website.
\newblock \emph{Tobacco control}, 16\penalty0 (3):\penalty0 207--210, 2007.

\bibitem[Snelson(2015)]{snelson2015vlogging}
Chareen Snelson.
\newblock Vlogging about school on youtube: An exploratory study.
\newblock \emph{New Media \& Society}, 17\penalty0 (3):\penalty0 321--339, 2015.

\bibitem[Huh et~al.(2014)Huh, Liu, Neogi, Inkpen, and Pratt]{huh2014health}
Jina Huh, Leslie~S Liu, Tina Neogi, Kori Inkpen, and Wanda Pratt.
\newblock Health vlogs as social support for chronic illness management.
\newblock \emph{ACM Transactions on Computer-Human Interaction (TOCHI)}, 21\penalty0 (4):\penalty0 1--31, 2014.

\bibitem[Gratch et~al.(2014)Gratch, Artstein, Lucas, Stratou, Scherer, Nazarian, Wood, Boberg, DeVault, Marsella, et~al.]{gratch2014distress}
Jonathan Gratch, Ron Artstein, Gale Lucas, Giota Stratou, Stefan Scherer, Angela Nazarian, Rachel Wood, Jill Boberg, David DeVault, Stacy Marsella, et~al.
\newblock The distress analysis interview corpus of human and computer interviews.
\newblock Technical report, University of Southern California Los Angeles, 2014.

\bibitem[Girard et~al.(2013)Girard, Cohn, Mahoor, Mavadati, and Rosenwald]{girard2013social}
Jeffrey~M Girard, Jeffrey~F Cohn, Mohammad~H Mahoor, Seyedmohammad Mavadati, and Dean~P Rosenwald.
\newblock Social risk and depression: Evidence from manual and automatic facial expression analysis.
\newblock In \emph{Proc. of FG '13}, pages 1--8. IEEE, 2013.

\bibitem[Ray et~al.(2019)Ray, Kumar, Reddy, Mukherjee, and Garg]{ray2019multi}
Anupama Ray, Siddharth Kumar, Rutvik Reddy, Prerana Mukherjee, and Ritu Garg.
\newblock Multi-level attention network using text, audio and video for depression prediction.
\newblock In \emph{Proc. of AVEC '19}, pages 81--88, 2019.

\bibitem[Joshi et~al.(2013)Joshi, Dhall, Goecke, and Cohn]{joshi2013relative}
Jyoti Joshi, Abhinav Dhall, Roland Goecke, and Jeffrey~F Cohn.
\newblock Relative body parts movement for automatic depression analysis.
\newblock In \emph{2013 Humaine association conference on affective computing and intelligent interaction}, pages 492--497. IEEE, 2013.

\bibitem[Conus et~al.(2014)Conus, Macneil, and McGorry]{conus2014public}
Philippe Conus, Craig Macneil, and Patrick~D McGorry.
\newblock Public health significance of bipolar disorder: implications for early intervention and prevention.
\newblock \emph{Bipolar disorders}, 16\penalty0 (5):\penalty0 548--556, 2014.

\bibitem[Thieme et~al.(2020)Thieme, Belgrave, and Doherty]{thieme2020machine}
Anja Thieme, Danielle Belgrave, and Gavin Doherty.
\newblock Machine learning in mental health: A systematic review of the hci literature to support the development of effective and implementable ml systems.
\newblock \emph{ACM Transactions on Computer-Human Interaction (TOCHI)}, 27\penalty0 (5):\penalty0 1--53, 2020.

\bibitem[Hao et~al.(2013)Hao, Li, Li, and Zhu]{hao2013predicting}
Bibo Hao, Lin Li, Ang Li, and Tingshao Zhu.
\newblock Predicting mental health status on social media: A preliminary study on microblog.
\newblock In \emph{Proc. of CCD '13}, pages 101--110. Springer, 2013.

\bibitem[Wang et~al.(2017)Wang, Brede, Ianni, and Mentzakis]{wang2017detecting}
Tao Wang, Markus Brede, Antonella Ianni, and Emmanouil Mentzakis.
\newblock Detecting and characterizing eating-disorder communities on social media.
\newblock In \emph{Proc. of WSDM '17}, pages 91--100, 2017.

\bibitem[Halfin(2007)]{halfin2007depression}
Aron Halfin.
\newblock Depression: the benefits of early and appropriate treatment.
\newblock \emph{American Journal of Managed Care}, 13\penalty0 (4):\penalty0 S92, 2007.

\bibitem[Xu et~al.(2019)Xu, Zhang, Li, Li, and Yip]{xu2019individualized}
Zhongzhi Xu, Qingpeng Zhang, Wentian Li, Mingyang Li, and Paul Siu~Fai Yip.
\newblock Individualized prediction of depressive disorder in the elderly: a multitask deep learning approach.
\newblock \emph{International journal of medical informatics}, 132:\penalty0 103973, 2019.

\bibitem[King et~al.(2008)King, Walker, Levy, Bottomley, Royston, Weich, Bellon-Saameno, Moreno, {\v{S}}vab, Rotar, et~al.]{king2008development}
Michael King, Carl Walker, Gus Levy, Christian Bottomley, Patrick Royston, Scott Weich, Juan~Angel Bellon-Saameno, Berta Moreno, Igor {\v{S}}vab, Danica Rotar, et~al.
\newblock Development and validation of an international risk prediction algorithm for episodes of major depression in general practice attendees: the predictd study.
\newblock \emph{Archives of general psychiatry}, 65\penalty0 (12):\penalty0 1368--1376, 2008.

\bibitem[De~Choudhury et~al.(2013)De~Choudhury, Gamon, Counts, and Horvitz]{de2013predicting}
Munmun De~Choudhury, Michael Gamon, Scott Counts, and Eric Horvitz.
\newblock Predicting depression via social media.
\newblock In \emph{Proc. of AAAI '13}, volume~7, pages 128--137, 2013.

\bibitem[De~Choudhury(2013)]{de2013role}
Munmun De~Choudhury.
\newblock Role of social media in tackling challenges in mental health.
\newblock In \emph{Proc. of SAM '13}, pages 49--52, 2013.

\bibitem[De~Choudhury(2015)]{de2015social}
Munmun De~Choudhury.
\newblock Social media for mental illness risk assessment, prevention and support.
\newblock In \emph{Proc. of SIdEWayS `15}, page~1, 2015.

\bibitem[Balani and De~Choudhury(2015)]{balani2015detecting}
Sairam Balani and Munmun De~Choudhury.
\newblock Detecting and characterizing mental health related self-disclosure in social media.
\newblock In \emph{Proc. of CHI EA '15}, pages 1373--1378, 2015.

\bibitem[Al~Asad et~al.(2019)Al~Asad, Pranto, Afreen, and Islam]{al2019depression}
Nafiz Al~Asad, Md~Appel~Mahmud Pranto, Sadia Afreen, and Md~Maynul Islam.
\newblock Depression detection by analyzing social media posts of user.
\newblock In \emph{Proc. of SPICSCON '19}, pages 13--17. IEEE, 2019.

\bibitem[Pirina and {\c{C}}{\"o}ltekin(2018)]{pirina2018identifying}
Inna Pirina and {\c{C}}a{\u{g}}r{\i} {\c{C}}{\"o}ltekin.
\newblock Identifying depression on reddit: The effect of training data.
\newblock In \emph{Proc. of the 2018 EMNLP Workshop SMM4H: The 3rd Social Media Mining for Health Applications Workshop \& Shared Task}, pages 9--12, 2018.

\bibitem[Tadesse et~al.(2019)Tadesse, Lin, Xu, and Yang]{tadesse2019detection}
Michael~M Tadesse, Hongfei Lin, Bo~Xu, and Liang Yang.
\newblock Detection of depression-related posts in reddit social media forum.
\newblock \emph{IEEE Access}, 7:\penalty0 44883--44893, 2019.

\bibitem[Guo et~al.(2022)Guo, Zhu, Hao, and Hong]{guo2022automatic}
Yanrong Guo, Chenyang Zhu, Shijie Hao, and Richang Hong.
\newblock Automatic depression detection via learning and fusing features from visual cues.
\newblock \emph{IEEE Transactions on Computational Social Systems}, 2022.

\bibitem[Wang et~al.(2018)Wang, Yang, and Yu]{wang2018facial}
Qingxiang Wang, Huanxin Yang, and Yanhong Yu.
\newblock Facial expression video analysis for depression detection in chinese patients.
\newblock \emph{Journal of Visual Communication and Image Representation}, 57:\penalty0 228--233, 2018.

\bibitem[Chen et~al.(2021)Chen, Chaturvedi, Ji, and Cambria]{chen2021sequential}
Qian Chen, Iti Chaturvedi, Shaoxiong Ji, and Erik Cambria.
\newblock Sequential fusion of facial appearance and dynamics for depression recognition.
\newblock \emph{Pattern Recognition Letters}, 150:\penalty0 115--121, 2021.

\bibitem[Jocher et~al.(2020)Jocher, Stoken, Borovec, NanoCode012, ChristopherSTAN, Changyu, Laughing, tkianai, Hogan, lorenzomammana, yxNONG, AlexWang1900, Diaconu, Marc, wanghaoyang0106, ml5ah, Doug, Ingham, Frederik, Guilhen, Hatovix, Poznanski, Fang, 于力军, changyu98, Wang, Gupta, Akhtar, PetrDvoracek, and Rai]{glenn_jocher_2020_4154370}
Glenn Jocher, Alex Stoken, Jirka Borovec, NanoCode012, ChristopherSTAN, Liu Changyu, Laughing, tkianai, Adam Hogan, lorenzomammana, yxNONG, AlexWang1900, Laurentiu Diaconu, Marc, wanghaoyang0106, ml5ah, Doug, Francisco Ingham, Frederik, Guilhen, Hatovix, Jake Poznanski, Jiacong Fang, Lijun~Yu 于力军, changyu98, Mingyu Wang, Naman Gupta, Osama Akhtar, PetrDvoracek, and Prashant Rai.
\newblock {ultralytics/yolov5: v3.1 - Bug Fixes and Performance Improvements}.
\newblock \url{https://doi.org/10.5281/zenodo.4154370}, October 2020.
\newblock Accessed: 2023-06-08.

\bibitem[Kipf and Welling(2016)]{kipf2016semi}
Thomas~N Kipf and Max Welling.
\newblock Semi-supervised classification with graph convolutional networks, 2016.

\bibitem[Hamilton et~al.(2017)Hamilton, Ying, and Leskovec]{hamilton2017inductive}
Will Hamilton, Zhitao Ying, and Jure Leskovec.
\newblock Inductive representation learning on large graphs.
\newblock \emph{Advances in neural information processing systems}, 30, 2017.

\bibitem[Veli{\v{c}}kovi{\'c} et~al.(2018)Veli{\v{c}}kovi{\'c}, Cucurull, Casanova, Romero, Li{\`o}, and Bengio]{velivckovic2017graph}
Petar Veli{\v{c}}kovi{\'c}, Guillem Cucurull, Arantxa Casanova, Adriana Romero, Pietro Li{\`o}, and Yoshua Bengio.
\newblock Graph attention networks, 2018.

\bibitem[He et~al.(2016)He, Zhang, Ren, and Sun]{he2016deep}
Kaiming He, Xiangyu Zhang, Shaoqing Ren, and Jian Sun.
\newblock Deep residual learning for image recognition.
\newblock In \emph{Proc. of CVPR '16}, pages 770--778, 2016.

\bibitem[{SKTBrain}(2023)]{kobert}
{SKTBrain}.
\newblock Kobert.
\newblock \url{https://github.com/SKTBrain/KoBERT}, 2023.
\newblock 2023-06-08.

\bibitem[Pires et~al.(2019)Pires, Schlinger, and Garrette]{pires-etal-2019-multilingual}
Telmo Pires, Eva Schlinger, and Dan Garrette.
\newblock How multilingual is multilingual {BERT}?
\newblock In \emph{Proc. of ACL '19}, Florence, Italy, July 2019. Association for Computational Linguistics.
\newblock \doi{10.18653/v1/P19-1493}.
\newblock URL \url{https://aclanthology.org/P19-1493}.

\bibitem[Lample and Conneau(2019)]{lample2019cross}
Guillaume Lample and Alexis Conneau.
\newblock Cross-lingual language model pretraining, 2019.

\bibitem[Vaswani et~al.(2017)Vaswani, Shazeer, Parmar, Uszkoreit, Jones, Gomez, Kaiser, and Polosukhin]{vaswani2017attention}
Ashish Vaswani, Noam Shazeer, Niki Parmar, Jakob Uszkoreit, Llion Jones, Aidan~N Gomez, {\L}ukasz Kaiser, and Illia Polosukhin.
\newblock Attention is all you need.
\newblock \emph{Advances in neural information processing systems}, 30, 2017.

\bibitem[Lin et~al.(2014)Lin, Maire, Belongie, Hays, Perona, Ramanan, Doll{\'a}r, and Zitnick]{lin2014microsoft}
Tsung-Yi Lin, Michael Maire, Serge Belongie, James Hays, Pietro Perona, Deva Ramanan, Piotr Doll{\'a}r, and C~Lawrence Zitnick.
\newblock Microsoft coco: Common objects in context.
\newblock In \emph{Proc. of ECCV '14}, pages 740--755. Springer, 2014.

\bibitem[Kingma and Ba(2014)]{kingma2014adam}
Diederik~P Kingma and Jimmy Ba.
\newblock Adam: A method for stochastic optimization, 2014.

\end{thebibliography}






\end{document}